# A Choquet Fuzzy Integral Vertical Bagging Classifier for Mobile Telematics Data Analysis

Mohammad Siami, Mohsen Naderpour, Jie Lu
Decision Systems and e-Service Intelligence Laboratory
Centre for Artificial Intelligence (CAI), Faculty of Engineering and IT
University of Technology Sydney (UTS)
Ultimo, NSW 2007, Australia
Mohammad.SiamiNamini@student.uts.edu.au; Mohsen.Naderpour, Jie.Lu @uts.edu.au

*Abstract* — **Mobile app development in recent years has resulted in new products and features to improve human life. Mobile telematics is one such development that encompasses multidisciplinary fields for transportation safety. The application of mobile telematics has been explored in many areas, such as insurance and road safety. However, to the best of our knowledge, its application in gender detection has not been explored. This paper proposes a Choquet fuzzy integral vertical bagging classifier that detects gender through mobile telematics. In this model, different random forest classifiers are trained by randomly generated features with rough set theory, and the top three classifiers are fused using the Choquet fuzzy integral. The model is implemented and evaluated on a real dataset. The empirical results indicate that the Choquet fuzzy integral vertical bagging classifier outperforms other classifiers.**

## I. Introduction

Recent technological improvements in smartphones and mobile app development have provided new capabilities for users. Mobile telematics is one of the newest advances; it records users' driving characteristics through the internal hardware of mobile phones instead of via built-in vehicle data recorders. In-vehicle data recorders are very expensive, and in recent years there has been a tendency to use mobile telematics instead of vehicle telematics. This monitoring technology uses embedded velocity and acceleration sensors together with the Global Positioning System (GPS) in mobile hardware to transmit driving characteristics to an external server, which stores the data for future analytics. This technology, invented by Malalur et al. [1], records such driving characteristics as vehicle movements, acceleration, velocity and changes in these parameters under various conditions. The large quantity of streamed data captured by these devices provides an accurate picture of a driver's habits [2]. However, the captured data is very big and brings with it all the benefits and impediments of big data; it is highly valuable but also very challenging due to its volume, velocity, variety, and veracity.

Mobile telematics has so far been proposed in a number of road safety applications [3], intelligent transportation systems [4], and usage-based insurance [5], but applying this technology in real-world businesses is problematic. One challenge is driver identification, for which Dong et al. [6], proposed an artificial intelligence algorithm as the solution. According to the opinion of experts in Usage-Based insurance, missing data is another challenge in the application of mobile telematics, because most users prefer not to complete their demographic information in the mobile app. The gender of users is one of these unknown features. To the best of our knowledge, mobile telematics has so far been unable to reliably answer this question: what is the gender of the driver behind the wheel, male or female? This raises the following research questions, which form the focus of this study. 1) Is there any relation between driving habits and a driver's gender? 2) How can we apply driving stream data in machine learning algorithms? 3) What are the characteristics of a machine learning algorithm capable of achieving the highest performance in gender detection in the driving data domain?

To answer these questions, we propose a new framework for Mobile telematics data analytics based on the driving pattern of the driver. In addition to this framework, the paper makes further contributions by proposing a feature extraction methodology to transform unstructured driving data into features that are understandable by machines, and introduces a novel Choquet fuzzy integral vertical bagging algorithm. The proposal is implemented and evaluated in a Python 3 framework using the sklearn machine learning library and through a driving characteristics dataset of 408 trips made by 301 unique drivers, collected from a Usage-Based insurance company.

The rest of this paper is organized as follows. Section II provides a brief overview of driving style analytics, gender detection, and fuzzy integral. Section III details the proposed methodology. Section IV presents the experimental results. Section V concludes the paper and describes future work.

## II. Literature Review

### A. Driving Style Analytics and Mobile Telematics

Mobile telematics is a new technology, but driving style behavioral data analytics is not new. Meiring and Myburgh [7] provided a literature review of different driving styles. They categorized driving styles into four main groups: safe, aggressive, inattentive, and drunk. They also investigated different applications that assess driving behavior for driving assistance, fleet management, drowsiness detection, and insurance application. In recent years, an increasing amount of literature has applied machine learning to driving style analysis. Wang and Xi [8] proposed a binary classification solution to distinguish between aggressive and moderate driving patterns. They proposed a k-means clustering-based support vector machine method (kMC-SVM) to decrease the execution time and improve classification performance. In another study,

Henriksson [9] developed a pattern recognition framework to classify the driving context according to the data generated by the vehicle. The author differentiated between city driving style and highway driving style by finding the hidden relation between driving attributes on city roads or highways. He used four classification algorithms, Logistic Regression, SVM, Hidden Markov Model and a simple Baseline model. The results indicate that SVM achieves the best performance compared to the other methods.

Gender detection is generally considered to mean the identification of a person's gender by understanding certain male or female characteristics. There are many physical, physiological, and behavioral differences between males and females, as demonstrated by the results of various surveys [10, 11]. These differences are easily understandable by humans, who can often recognize the gender of persons by looking at their face or listening to their voice, but the task is very difficult for automated machines [12]. In recent years, many algorithms have been developed to automate gender identification from a range of data sources including speech [13], facial image [14], hand-writing [15], and biomedical data [12], but to the best of our knowledge, developing an automated system for gender recognition from driving style data is still an open issue.

*B. Fuzzy Integral*

Classification results are not usually precise or certain, so fuzzy theory is useful for merging different classifiers into one prediction result. It has been shown that fuzzy integral methods such as Sugeno and Choquet are popular and practical methods that have been used in a wide range of domains including mathematics, economics, machine learning and pattern recognition [16].

Although both of these integral methods are fuzzy and popular, Choquet fuzzy integrals have been more widely applied than Sugeno integrals [17]. A Choquet integral is an aggregation method that simultaneously considers the importance of a classifier and its interaction with other classifiers [18]. It relies on the concept of fuzzy measures first introduced by Sugeno [19]. The definitions of Choquet integrals and fuzzy measures according to [20] are as follows.

Assume X is a set of classifiers and the power of X is denoted by P(X).

<u>Definition 1:</u> The fuzzy measure of X is a set function $g: P(X) \rightarrow [0,1]$. This function satisfies the following conditions:

1) The Boundary of g is : $g(\phi) = 0, g(X) = 1$
2) For each $A, B \in P(X)$ and $A \subset B$ then $g(A) \leq g(B)$

where $g(k)$ is the grade of subjective importance of the classifier set $k$. The fuzzy singleton measure values for each classifier are $g(x_i) = g^i$ and are commonly called densities. Not only must the worth of each singleton be calculated, but also the value of function g for any combination of classifiers. The Sugeno λ-measure and fuzzy densities are used to calculate the fuzzy measure of any combination of classifiers. This measure is defined by the values of the fuzzy densities. The λ-measure has the following additional property:

$$\begin{cases} g_\lambda(A \cup B) = g_\lambda(A) + g_\lambda(B) + \lambda g_\lambda(A)g_\lambda(B) \\ \forall A, B \in P(X), A \cap B = \phi \end{cases} \quad (1)$$

where $\lambda$ can be calculated by Eq. 2.

$$\lambda + 1 = \prod_{i=1}^{n}(1 + \lambda g^i), \lambda > -1 \quad (2)$$

<u>Definition 2:</u> g is the fuzzy measure of $X = \{x_1, x_2, ..., x_n\}$. Eq. 3 shows the Choquet integral function of $f: X \rightarrow R$ and its relation to g:

$$C_g(f) = \sum_{i=1}^{n} f_i \, [\, g(A_i) - g(A_{i-1})] \quad (3)$$

A permutation of X is indicated by $(i)$, and $f(x_{(1)}) \leq f(x_{(2)}) \leq \cdots \leq f(x_{(n)})$ also $A_i = \{x_{(i)}, x_{(i+1)}, ..., x_{(n)}\}, A_0 = \phi$.

The prediction result of classifier $x_i$, is denoted by $f_i$, and $[g(A_i) - g(A_{i-1})]$ depicts the relative importance of the classifier $x_i$. The fuzzy integral of $f$ with respect to $g$ is the integration result.

III. METHODOLOGY

The methodology is illustrated in Fig. 1 and explained in the following sub-sections.

*A. Driving Style Dataset*

Table I introduces the driving characteristics stream data used in this research. These data were generated by mobile telematics at a sample rate of 15 samples per second (15 Hz).

TABLE I. STREAM DATA INTRODUCTION

| Name of feature | Description |
|---|---|
| Speed | The value of the instantaneous velocity of the vehicle stored by a smartphone |
| Acceleration (x,y) | Two different stream data of a car's acceleration over x and y axis. |
| Yaw rate | The value of the vehicle's angular speed around its vertical axis. |
| Pitch rate | The lateral motion of a car is called the pitch rate. The pitch rate value shows the up or down forward tilt of the vehicle. |
| Roll rate | The longitudinal axis movement of the car shows the characteristics of the road. |
| GPS heading | The compass direction measured in degrees from North. |

*B. Data Pre-Processing*

The first step for any big data analytics project is data preparation or data pre-processing. The data generated by smartphones are unstructured and are not comprehensible by machine learning algorithms. It is therefore necessary before starting analytics to transform these data into a new form of data by cleaning them, removing outliers, and extracting and selecting features. We considered previous research on human activity recognition for feature extraction that uses smartphone data similar in structure to our research dataset [21, 22].

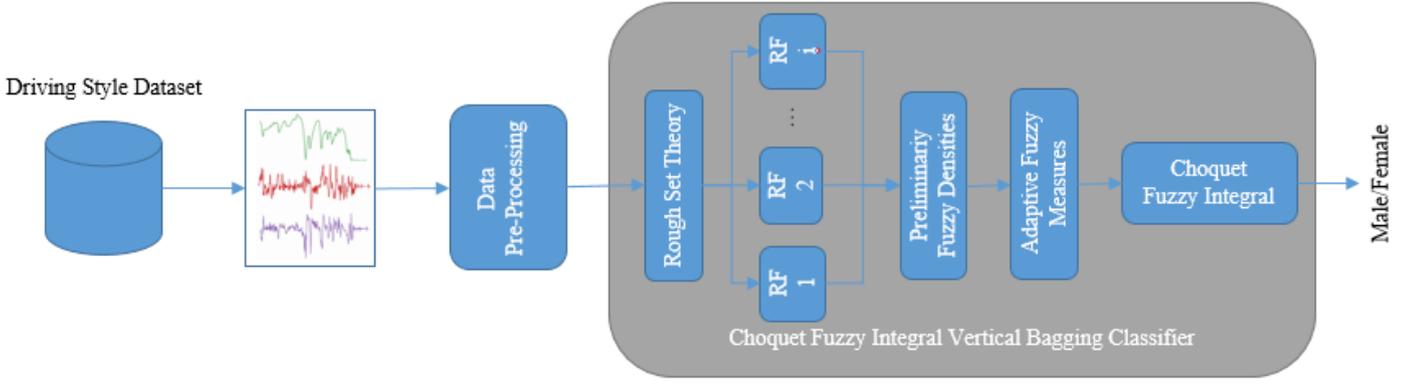

Fig. 1- Choquet Fuzzy Integral Vertical Bagging Classifier

The sample rate of the input data is equal to 15 Hz. This sample rate is high for our methodology, thus we down-sampled the data to one sample per second because our investigations revealed that one sample per second produces the best information for our analytics. We developed a windowing procedure to select a driving time window of length of 256 and sliding size 256. We summarized the data to develop statistical features from stream data for each time window. We calculated 14 statistical features including minimum, maximum, mean, median, first and third quantile, standard deviation, average absolute deviation, skewness, entropy, kurtosis, auto-correlation, zero crossing, energy for each stream data feature. We calculated these features for all seven stream data, extracting 7×14 features for each time window.

The next step in data pre-processing is feature selection. We removed features with low variation and, by developing a correlation analysis, we determined the correlation between all the extracted features and the gender of a driver, which we then used to select the most important features and remove useless attributes. After finalizing the data pre-processing, our data were clean and ready for the analytical tasks.

## C. Choquet Fuzzy Integral Vertical Bagging Classifier

The vertical bagging model is similar to traditional bagging, the difference being in the method of creating sub-models. In traditional bagging, sub-models are generated from sub-samples of data with the same attributes. In contrast, the sub-models in vertical bagging models are trained by various combinations of predictive attributes with similar sample data. [23]. In this paper, we combine multiple random forest classifiers with Choquet fuzzy integral to propose the Choquet fuzzy integral vertical bagging classifier. The maximum number of features (F) and the maximum number of iterations are the input parameters for training the algorithm.

Using rough set theory, k training subsets $Tr_1$, $Tr_2$…. $Tr_k$ were generated from the input training data. Maximum number of features (F) is equal to the total number of features in each Tr subset and k is equal to the maximum number of iterations. Random Forest (RF) classifiers were trained for each training subset. All RF models were sorted according to their performance in training dataset, and the top three models were selected. In the next step, the top random forest classifiers were merged by using Choquet fuzzy integral, which achieves outstanding performance in merging different classifiers [24].

*a) Preliminary Fuzzy Densities:*

In the previous step, the top three random forest classifiers with the highest performance in the training data were selected. The confusion matrix for each classifier was determined. Eq. 3 was used to calculate the Choquet fuzzy integral on the fuzzy measures (g), reflecting the importance of each classifier and classifier combinations. Equation 1 was used to calculate the value of any combination of classifiers. These parameters are crucial in the fuzzy integral vertical bagging classifier, and they play an essential role in the practical application of fuzzy integral in this algorithm and information fusion.

The confusion matrix of the i-th classifier is defined as

$$CM_i = \begin{bmatrix} n^i_{11} & \cdots & n^i_{1M} \\ \vdots & \ddots & \vdots \\ n^i_{M1} & \cdots & n^i_{MM} \end{bmatrix} \quad i = 1,2,\ldots,P \quad (4)$$

where $j_1 = j_2$, $n^i_{j_1 j_2}$ depicts the number of samples in class $c_{j1}$ which are correctly classified as $c_{j1}$ by the i-th classifier. On the other hand, $j_1 \neq j_2$, $n^i_{j_1 j_2}$ represents the number of samples with class label $c_{j1}$, but they have been misclassified as $c_{j2}$ by the classifier i. Therefore, the probability of items which have been correctly classified is calculated by Eq. 5 for each classifier and class label.

$$p^i_{j_1 j} = p(s_k \in c_{j1} | E_i(s_k) = c_j) = \frac{n^i_{j_1 j}}{\sum_{j=1}^{M} n^i_{j_1 j}} \quad (5)$$

$(j_1 = 1,2,\ldots,M; j = 1,2,\ldots,M)$

where i is the i-th classifier, M is the number of classes, and the probability matrix is

$$PM_i = \begin{bmatrix} p^i_{11} & \cdots & p^i_{1M} \\ \vdots & \ddots & \vdots \\ p^i_{M1} & \cdots & p^i_{MM} \end{bmatrix} \quad i = 1,2,\ldots,P \quad (6)$$

The $p^i_{jj}$ elements in $PM_i$ represent the percentage of items which are classified correctly by the $E_i$ classifier. Let $g^i_j = p^i_{jj}$, then $g_j^i$ depicts the preliminary fuzzy density value for the j-th class with respect to the i-th classifier. The fuzzy density value of each classifier and the different class labels is the output of this step.

*b) Adaptive Fuzzy Measures:*

The classification results of random forest differ according to the feature set. Some classifiers may have better performance than others, and one classifier may be more robust than others for classifying certain types of items or classes. Merging classifiers by voting strategy or by assigning equal value to all classifiers is therefore not an efficient approach, and the fuzzy density measure ($g_j^i$) needs to be properly adjusted by considering all classifiers and all classes.

After the correct classification rates and misclassification errors within the classifiers have been calculated, the values are used to update the fuzzy densities. The fuzzy densities are then updated by considering the pairwise proportion of wrongly classified items between the selected classifiers and others. The fuzzy density parameters can be updated by using Eq. 7 [25]:

$$g_j^{*i} = g_j^i * \left(\prod_m \delta_j^{i/m}\right)^{w1} * \left(\prod_m \gamma_j^{i/m}\right)^{w2} \quad (7)$$

where $g_j^{*i}$ is the updated fuzzy density for the i-th classifier for class j; $\{\delta_j^{i/m}\}, 0 < \delta_j^{i/m} < 1$, and $\{\gamma_j^{i/m}\}, 0 < \gamma_j^{i/m} < 1$ are the sets of updated parameters. $g_j^{*i}$ is calculated for all classes. Each set of updated parameters could have a different impact on the final outcome, so to add flexibility to the final result, $w_1$ and $w_2$ in Eq. 7 are used in updating process.

$\delta^{i/m}$ is used to update the initial fuzzy density when the output of two different classifiers do not have the same result. The initial fuzzy density of classifier will be decreased by increasing the number of misclassified objects, while the correctly classified items will increase the power of the classifier by using Eq. 8.

$$\delta_j^{i/m} = f(x) = \begin{cases} 1 & ,k_i/i = k_2/m \\ \dfrac{p_{j/i,j/i}^i - p_{k/i,j/m}^i}{p_{j/i,j/i}^i} & ,k_1/i \neq k_2/m \end{cases} \quad (8)$$

where $k_1$/i, and $k_2$/m show that class $k_1$ is given by classifier $E_i$, and class $k_2$ is given by classifier $E_m$. When $k_1$/i $= k_2$/m , means that two classifiers have identified samples in similar classes. $k_1$/i $\neq k_2$/m means that two different classifiers have categorized a sample into two different classes. One sample may be correctly classified by $E_i$ in class C1, but misclassified by classifier $E_m$. The proportion of objects correctly classified by $E_i$ for class j is depicted by $P_{j/i,j/i}^i$, but the number of correctly classified items by other classifiers is $p_{k/i,j/m}^i$. The training dataset is used to obtain both $P_{j/i,j/i}^i$ and $p_{k/i,j/m}^i$. Once the number of items misclassified by $E_i$ have increased, the corresponding fuzzy density measure of the $E_i$ classifier will be decreased.

The reason for updating the parameters $\gamma_j^{i/m}$ is that the initial fuzzy density of a classifier should be reduced when the error $E_i$ is more than the classifier $E_m$, but the fuzzy density value is not changed if the classifier $E_i$ has the same or fewer mistakes than classifier $E_m$. This concept is developed by Eq. 9.

$$\gamma_j^{i/m} = \begin{cases} 1 & : p_{k/i,q/m}^i \leq p_{k/i,q/m}^m \\ \dfrac{p_{k/i,q/m}^m}{p_{k/i,q/m}^i} & : p_{k/i,q/m}^i > p_{k/i,q/m}^m \\ \varepsilon & : p_{k/i,q/m}^m = 0 \end{cases} \quad (9)$$

where $\varepsilon$ is a very small value, which prevents $\gamma_j^{i/m}$ from being zero.

Adjusted fuzzy density is the output of this step, which updates the importance of each classifier in the training dataset by considering its performance in training dataset classifying items correctly or misclassifying them.

*c) Choquet Fuzzy Integral:*

The performance of each classifier in the vertical bagging RF is variant. Some RF classifiers have insufficient power to predict the result correctly, while another RF model may achieve excellent performance on the same samples. We propose the Choquet fuzzy integral vertical bagging random forest to take advantage of different random forest models. Choquet fuzzy integral fuses the results of multiple classifiers and provides a robust classifier with a more consistent result.

Suppose in a sample data space S, data is divided into two classes by a classifier (E). A classifier index is specified by ($i = 1, ..., P$); j is the class index ($j = 1, ..., M$); and k is the instance index($k = 1, ..., N$). For k-th sample the prediction result by the i-th classifier is $[h_{i1}(k), h_{i2}(k), ..., h_{iM}(k)]$ where $h_{ij}(k)$ is the probability result of the ith classifier, which shows the probability of k-th data belonging to class j. $[h_{1j}(k), h_{2j}(k), ..., h_{Pj}(k)]^T$ is defined as $h_j(s_k)$ which can be interpreted as :

$h_j: S \rightarrow [0,1]$ , $h_j(s_k) = [h_{1j}(k), h_{2j}(k), ..., h_{Pj}(k)]^T$ for sample $s_k$, we obtain a value for $h_j(s_k)$ as degree of support provided by each classifier with respect to the j-th class for sample $s_k$.

In addition to $h_j(s_k)$, the Choquet fuzzy integral operates on the fuzzy measures (g). Fuzzy measures include fuzzy densities and the fuzzy measure of any combination of classifiers, which are calculated in Eqs. 8 and 1.

By calculating the Choquet integral of $h_j(s_k)$, g, we can provide the degree of support given by the ensemble classifier with respect to the j-th class for sample $s_k$. The output class $c_j$ for the sample $s_k$ is the class with the largest integral value:

$$c_j = arg(\max_{1 \leq l \leq M} \int h_l(s_k) dg) \quad (10)$$

A summary of the Choquet fuzzy integral vertical bagging classifier is as follows:

```
Input: Data, maximum number of features (F),
maximum iteration (K)
  1) Generate a list of important features in feature
     engineering step
  2) For k in [1,2, … K] as iteration:
     ✓ Train RF with F number of random features
     ✓ Validate RF with training dataset
     ✓ End for k
  3) Select top three RF models for fusion with
     training dataset
  4) Construct the confusion matrix for each selected
     classifier (Eq. 4), with training dataset
  5) For each j in [1,2] as [male/female]:
     ✓ For each i in [1,2,3] as top RF classifiers:
       • Calculate initial fuzzy densities by Eq.5.
       • Update parameter [$\delta_j^{i/m}$] by Eq. 8.
       • Update parameter [$\gamma_j^{i/m}$] by Eq. 9.
       • Update the initial fuzzy densities by Eq. 7.
       • End for i
     ✓ Compute the $g_\lambda$- fuzzy measures with updated fuzzy
       densities
     ✓ Compute the fuzzy integral each class with Eq. 3.
     ✓ End for j
  6) Use Eq. 10 to detect the gender of a driver.
```

## IV. EXPERIMENTAL RESULTS

The primary goal of this section is to examine the prediction results of the methodology for gender detection from smartphone-generated data. We used almost 1 GB data containing the anonymized driving behavior of 301 unique drivers. These data were collected by a Usage-Based insurance company in real-world conditions. The dataset consists of streamed data of 408 trips. Each trip contains at least 15 minutes of driving data from 301 unique drivers, some of whom feature in more than one trip. The number of male drivers is 161 while the number of female drivers is 140. A brief description of the final dataset is summarized in Table II.

TABLE II. DATA DESCRIPTION

| Number of trips | Number of unique drivers | Total driving distance | Total driving time | Male | Female |
|---|---|---|---|---|---|
| 408 | 301 | 9898 km | 202 hours | 161 | 140 |

The experiment started by decreasing the stream data sample rate. We found that the best sample rate in our data for gender detection was one sample per second, which is the average of all 15 samples in one second. After decreasing the sample rate, we developed a windowing process. We segmented the driving characteristics into time windows which were equal to 512 seconds, then extracted all the proposed features listed in Section III for each time window. Features with very low variance were removed, and we developed a correlation analysis between the extracted features and gender of the driver to find the features of highest importance. We selected features which had a correlation greater than 0.1. The correlation analysis report is depicted in Table III.

After selecting the most valuable attributes from the extracted features, we had a clean data source ready for analytics containing 15 features of 2048 windows for 1119 male and 929 female. We used these data to validate our proposed classifier.

We developed a comparison analysis with three base classifiers: random forest, gradient boosting classifier, and logistic regression. Our model was trained by setting the maximum number of iterations to 100; the maximum number of features in rough set theory is 10. In addition to the input parameters for the vertical bagging classifier, we defined the exponential weights for $w_1$ and $w_2$ in Eq. 7 as $W_1$=0.9 , $W_2$=0.6 [24]. The value of ε in Eq. 9 was set to 0.0001.

TABLE III. CORRELATION ANALYSIS REPORT

| Feature name | Correlation |
|---|---|
| Speed mean | -0.19708638 |
| Speed_Q3 | -0.19611402 |
| Speed energy | -0.19610685 |
| Pitch_rate_Q1 | -0.19027125 |
| Speed median | -0.1869981 |
| Speed_Q1 | -0.18227661 |
| Pitch rate standard deviation | 0.17657519 |
| Speed max | -0.17616991 |
| Pitch rate energy | 0.1666721 |
| Pitch rate kurtosis | -0.16354984 |
| Pitch rate average absolute deviation | 0.1609966 |
| Pitch_rate_Q3 | 0.13283702 |
| Acceleration lon zero-crossing | -0.11766974 |
| Speed skewness | 0.11405972 |
| Yaw rate zero-crossing | -0.10055166 |

To evaluate the model's performance, we conducted 5-fold cross-validation. Accuracy, Area Under the Curve (AUC) are two performance measures in this research. AUC shows the area under the ROC curve, and the accuracy score reflects the proportion of correct items to all items. We have compared our model performance with three other classifiers, and the results are depicted in Table IV.

TABLE IV. COMPARISON OF RESULTS

| | AUC | Accuracy |
|---|---|---|
| **Our model** | **72.44** | **71.67** |
| Random Forest | 68.23 | 64.16 |
| Logistic Regression | 60.32 | 55.85 |
| Gradient Boosting Classifier | 66.16 | 62.51 |

The results in Table IV indicate that the Choquet fuzzy integral vertical bagging classifier achieves the best performance in accuracy and AUC compared to the selected alternative algorithms. These results show that the final model not only improves the performance of the base classifier, Random Forest, but that it also achieves better performance than Logistic Regression and Gradient Boosting.

In terms of accuracy, the proposed model has the best result for detecting gender. The accuracy score of the Choquet fuzzy integral vertical bagging classifier is 71.67, which is significantly higher than that of the classifiers selected for comparison. In addition to the accuracy score, we calculated the AUC score for each classifier. Our proposed model achieved 72.44, which is higher than all three comparison models. The logistic regression classifier returned the worst results for both accuracy and AUC score for gender detection compared to the other methods.

To gain a comprehensive view the performance of our model, and to prevent overfitting or underfitting, we conducted a 5-fold cross-validation. Fig. 2 shows the performance of all four models for each run. The results show that our model achieves the highest performance in all folds.

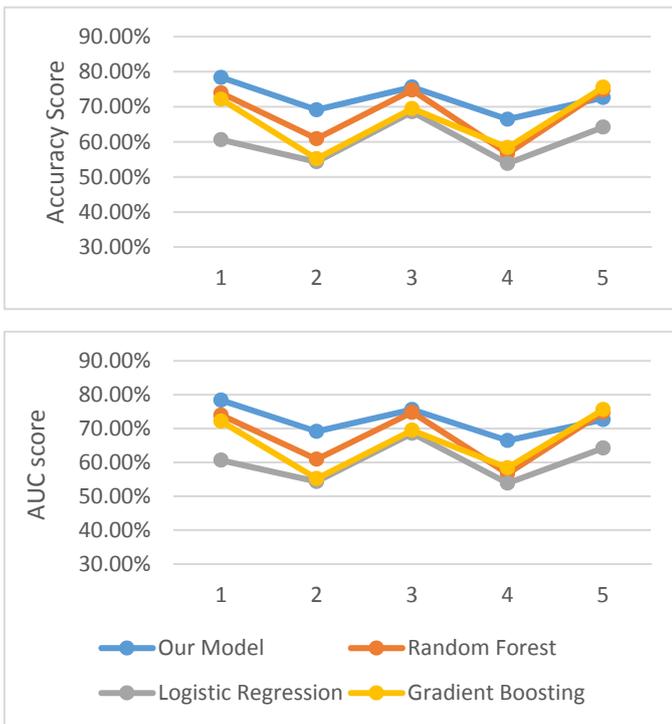

Fig. 2. Accuracy score and AUC score performance results on 5-fold

I. CONCLUSION AND FUTURE WORK

Technological improvements in mobile app development in recent years have offered new applications that can take advantage of smartphone capabilities. Recording driving habits is one such new application. In this paper, we have proposed a new Choquet fuzzy integral vertical bagging classifier for gender detection from smartphone-generated data. To the best of our knowledge, this is the first model to use smartphone data to detect the gender of a driver. The model uses randomly generated features to create several random forest. The Choquet fuzzy integral then aggregates the results of the top classifiers to find the final result. The empirical analysis shows that the vertical bagging classifier with Choquet fuzzy integral achieves the best accuracy and AUC score for gender detection compared to other classifiers. In future work, we will assess the performance of our model using other feature extraction methods. Moreover, we will assess the effects of variations in our model input variables, such as the maximum number of features and the maximum number of iterations, on the model's performance. Validating the model on other datasets is another goal of our future research.

REFERENCES

[1] P. G. Malalur, H. Balakrishnan, and S. R. Madden, "Telematics using personal mobile devices," ed: Google Patents, 2013.
[2] J. Wahlström, I. Skog, and P. Händel, "Driving behavior analysis for smartphone-based insurance telematics," in Proceedings of the 2nd Workshop on Physical Analytics, 2015, pp. 19-24.
[3] Y. Zhao, "Telematics: safe and fun driving," IEEE Intelligent Systems, vol. 17, pp. 10-14, 2002.
[4] Y. Zhao, "Mobile phone location determination and its impact on intelligent transportation systems," IEEE Transactions on Intelligent Transportation Systems, vol. 1, pp. 55-64, 2000.
[5] B. F. Bowne, N. R. Baker, D. L. Marzinzik, M. E. Riley, N. U. Christopulos, B. M. Fields, et al., "Methods to determine a vehicle insurance premium based on vehicle operation data collected via a mobile device," ed: Google Patents, 2013.
[6] W. Dong, J. Li, R. Yao, C. Li, T. Yuan, and L. Wang, "Characterizing driving styles with deep learning," arXiv preprint arXiv:1607.03611, 2016.
[7] G. A. M. Meiring and H. C. Myburgh, "A review of intelligent driving style analysis systems and related artificial intelligence algorithms," Sensors, vol. 15, pp. 30653-30682, 2015.
[8] W. Wang and J. Xi, "A rapid pattern-recognition method for driving styles using clustering-based support vector machines," in American Control Conference (ACC), 2016, pp. 5270-5275.
[9] M. Henriksson, "Driving context classification using pattern recognition," Master's Thesis, Chalmers University of Technology (University of Gothenburg), 2016.
[10] A. H. Eagly, Sex differences in social behavior: A social-role interpretation. Psychology Press, 2013.
[11] E. Mendoza, N. Valencia, J. Muñoz, and H. Trujillo, "Differences in voice quality between men and women: use of the long-term average spectrum (LTAS)," Journal of Voice, vol. 10, pp. 59-66, 1996.
[12] J. Hu, "An approach to EEG-based gender recognition using entropy measurement methods," Knowledge-Based Systems, vol. 140, pp. 134-141, 2018.
[13] H. Harb and L. Chen, "Voice-based gender identification in multimedia applications," Journal of Intelligent Information Systems, vol. 24, pp. 179-198, 2005.
[14] Z. Yang, M. Li, and H. Ai, "An experimental study on automatic face gender classification," in 18th International Conference on Pattern Recognition, ICPR 2006, pp. 1099-1102.
[15] M. Liwicki, A. Schlapbach, and H. Bunke, "Automatic gender detection using on-line and off-line information," Pattern Analysis and Applications, vol. 14, pp. 87-92, 2011.
[16] Q. Wang, C. Zheng, H. Yu, and D. Deng, "Integration of heterogeneous classifiers based on Choquet fuzzy integral," in 7th International Conference on Intelligent Human-Machine Systems and Cybernetics (IHMSC), 2015, pp. 543-547.
[17] A. R. Krishnan, M. M. Kasim, and E. M. N. E. A. Bakar, "A short survey on the usage of Choquet integral and its associated fuzzy measure in multiple attribute analysis," Procedia Computer Science, vol. 59, pp. 427-434, 2015.
[18] X. Li, F. Wang, and X. Chen, "Support vector machine ensemble based on Choquet integral for financial distress prediction," International Journal of Pattern Recognition and Artificial Intelligence, vol. 29, p. 1550016, 2015.
[19] M. Sugeno, "Theory of fuzzy integrals and its applications," Doctoral Thesis, Tokyo Institute of Technology, 1974.
[20] T. Murofushi and M. Sugeno, "An interpretation of fuzzy measures and the Choquet integral as an integral with respect to a fuzzy measure," Fuzzy Sets and Systems, vol. 29, pp. 201-227, 1989.
[21] O. D. Lara and M. A. Labrador, "A survey on human activity recognition using wearable sensors," IEEE Communications Surveys and Tutorials, vol. 15, pp. 1192-1209, 2013.
[22] M. M. Hassan, M. Z. Uddin, A. Mohamed, and A. Almogren, "A robust human activity recognition system using smartphone sensors and deep learning," Future Generation Computer Systems, vol. 81, pp. 307-313, 2018.
[23] D. Zhang, X. Zhou, S. C. Leung, and J. Zheng, "Vertical bagging decision trees model for credit scoring," Expert Systems with Applications, vol. 37, pp. 7838-7843, 2010.
[24] A. Namvar and M. Naderpour, "Handling uncertainty in social lending credit risk prediction with a Choquet fuzzy integral model," arXiv preprint arXiv:1804.10796, 2018.
[25] T. D. Pham, "Combination of multiple classifiers using adaptive fuzzy integral," in IEEE International Conference on Artificial Intelligence Systems, ICAIS 2002, pp. 50-55.